\documentclass[sigconf]{acmart}

\AtBeginDocument{%
  \providecommand\BibTeX{{%
    \normalfont B\kern-0.5em{\scshape i\kern-0.25em b}\kern-0.8em\TeX}}}

\copyrightyear{2022}
\acmYear{2022}
\setcopyright{acmcopyright}\acmConference[WWW '22]{Proceedings of the ACM Web
Conference 2022}{April 25--29, 2022}{Virtual Event, Lyon, France}
\acmBooktitle{Proceedings of the ACM Web Conference 2022 (WWW '22), April
25--29, 2022, Virtual Event, Lyon, France}
\acmPrice{15.00}
\acmDOI{10.1145/3485447.3511942}
\acmISBN{978-1-4503-9096-5/22/04}

\usepackage{float}
\newcommand{\mtt}[1]{\mbox{\tt #1}}

\newcommand{\temv}[1]{\mbox{\em #1}}

\begin{document}

\title{Unfreeze with Care: \\ Space-Efficient Fine-Tuning of Semantic Parsing Models}

\author{Weiqi Sun}
\affiliation{%
  \institution{Amazon Alexa AI}
  \city{New York}
  \country{USA}}
\email{weiqisun@amazon.com}

\author{Haidar Khan}
\affiliation{%
  \institution{Amazon Alexa AI}
  \city{New York}
  \country{USA}}
\email{khhaida@amazon.com}

\author{Nicolas Guenon des Mesnards}
\affiliation{%
  \institution{Amazon Alexa AI}
  \city{New York}
  \country{USA}}
\email{mesnarn@amazon.com}

\author{Melanie Rubino}
\affiliation{%
  \institution{Amazon Alexa AI}
  \city{New York}
  \country{USA}}
\email{rubinome@amazon.com}

\author{Konstantine Arkoudas}
\affiliation{%
  \institution{Amazon Alexa AI}
  \city{New York}
  \country{USA}}
\email{arkoudk@amazon.com}

\renewcommand{\shortauthors}{Sun and Khan, et al.}

\begin{abstract}
  Semantic parsing is a key NLP task that maps natural language to structured meaning representations. As in many other NLP tasks, SOTA performance in semantic parsing is now attained by fine-tuning a large pretrained language model (PLM). While effective, this approach is inefficient in the presence of multiple downstream tasks, as a new set of values for all parameters of the PLM needs to be stored for each task separately. Recent work has explored methods for adapting PLMs to downstream tasks while keeping most (or all) of their parameters frozen. We examine two such promising techniques, prefix tuning and bias-term tuning, specifically on semantic parsing. We compare them against each other on two different semantic parsing datasets, and we also compare them against full and partial fine-tuning, both in few-shot and conventional data settings. While prefix tuning is shown to do poorly for semantic parsing tasks off the shelf, we modify it by adding special token embeddings, which results in very strong performance without compromising parameter savings.
\end{abstract}





\maketitle

\section{Introduction}

Large transformer-based language models that have been pretrained without supervision on huge amounts of text have proven to be remarkably effective in powering downstream NLP tasks ranging from classification and regression to generation tasks such as question answering, summarization, and semantic parsing. Two common ways of transferring knowledge from a large pretrained language model (PLM) $M$ to a downstream task $T$ are: (a) using $M$ as a {\em feature extractor}, e.g., solely for the purpose of encoding a given input and passing the output to a subsequent model trained on $T$; and (b) \emph{fine-tuning} $M$ on $T$'s training data, after replacing $M$'s output layer by a task-specific head. Fine-tuning has become the de facto method of adapting a PLM to a downstream task, as most SOTA results in NLP have been obtained by this approach.
\begin{figure*}[h!]
    \centering
    \includegraphics[width=0.8\textwidth]{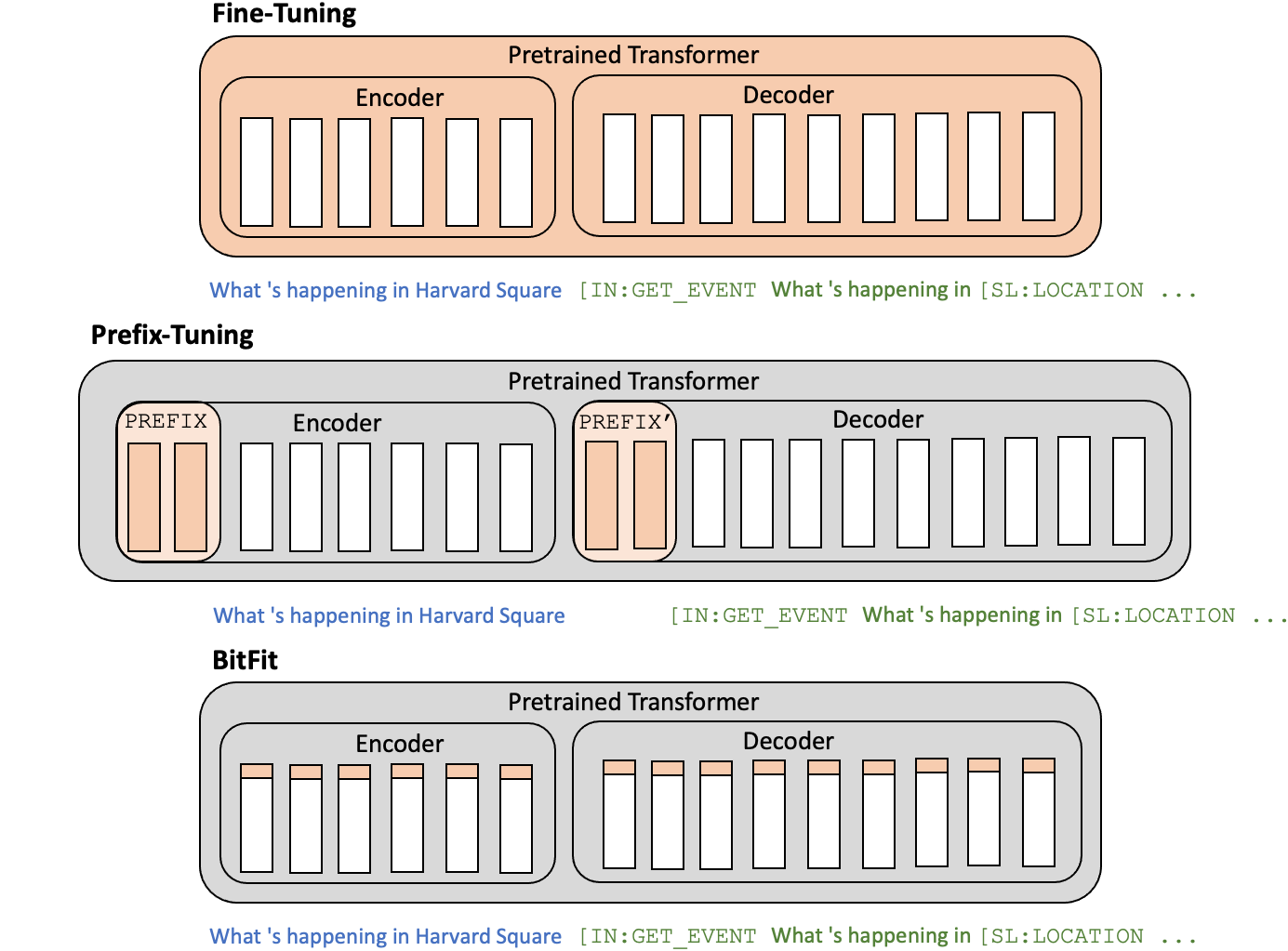}
    \caption{Fine-tuning updates all parameters.  Prefix-tuning freezes Transformer parameters and only optimizes the prefixes. BitFit freezes all the weight matrices and only updates the bias vector in the Transformer layers. White blocks are the hidden vectors, orange blocks represent the tunable model components, and grey blocks are frozen.}
    \label{fig:models}
\end{figure*}

While fine-tuning is very effective, it is susceptible to catastrophic forgetting unless the learning rate is carefully calibrated,\footnote{If the learning rate is too large then $M$ will forget what it learned during pre-training. On the other hand, if the learning rate is too small then $M$ will not adapt to $T$ well enough.}
 and can also be prone to overfitting in low-data settings. More importantly for present purposes, it requires updating and storing values for {\em all\/} of $M$'s parameters, which can easily number in the billions. In the presence of multiple downstream tasks (a frequent scenario in larger organizations), this approach becomes too inefficient. Accordingly, a good deal of effort has recently focused on adapting a PLM $M$ to downstream tasks in ways that do not require updating all of $M$'s parameters.

 We distinguish between \emph{mutative} methods of adapting $M$, which modify some or all of $M$'s parameters, and {\em non-mutative} methods, which {\em freeze\/} all of $M$ and instead learn additional parameters on a task-specific basis. Mutative methods do not need to modify all of $M$'s parameters. Indeed, it is common to modify only a few top layers of $M$. Non-mutative methods have the considerable advantage that multiple downstream tasks can share the same PLM, as they access it in a read-only way that does not change its weights. We introduce an additional distinction that partitions non-mutative methods into those that are \emph{opaque}, i.e., methods that do not need to know anything about the internal structure of $M$; and \emph{transparent} methods that are coupled with the specific architecture of $M$.

In this paper we study the effectiveness of two adaptation techniques specifically in connection with semantic parsing.
We focus on task-oriented parsing rather than the generation of general-purpose meaning representations (such as AMR \cite{banarescu-etal-2013-abstract}), although the techniques we discuss should be generally applicable. Semantic parsing (especially task-oriented parsing) is a key task for digital voice assistants such as Google Assistant and Alexa.
It also has a vital role to play in unlocking the full potential of the web, particularly in enabling sophisticated NLP-powered searches of web content.
Yet to the best of our knowledge there has been no prior investigation of adaptation techniques in the semantic parsing field. The two techniques that we investigate are prefix tuning, a transparent non-mutative method recently introduced by ~\citet{li2021prefixtuning}; and BitFit \cite{zaken2021bitfit}, a mutative technique that only fine-tunes bias terms; see Figure~\ref{fig:models}. We compare both approaches against full fine-tuning, as well as \emph{partial fine-tuning}, which only modifies the weights of a few top layers of the PLM. We study both regular- and low-resource data settings. 

\begin{figure}[b!]
\vspace*{-0.06in}
    \centering
    \includegraphics[width=0.48\textwidth]{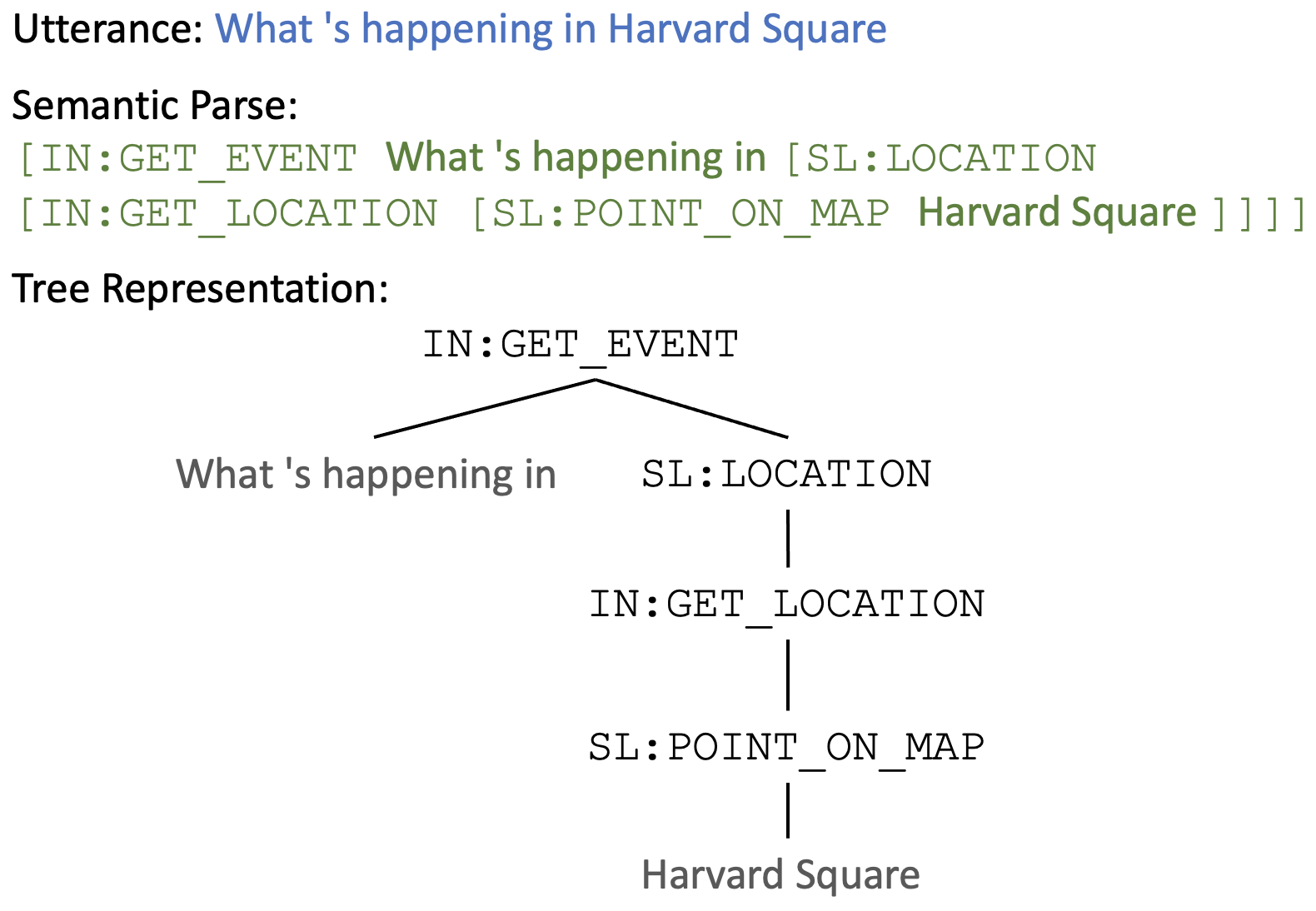}
\vspace*{-0.11in}
    \caption{A sample parse tree from the TOP dataset.}
    \label{fig:semparse}
\end{figure}

 We view semantic parsing as a sequence-to-sequence (seq2seq) task, where the input sequence $x$ is a natural-language utterance and the output sequence $y$ is a linearized semantic tree in some appropriate representation. An example from the TOP dataset \cite{gupta2018semantic} is shown in Figure~\ref{fig:semparse}. As described in Section~\ref{Sec:Datasets}, in this paper we work with the latest version of the TOP dataset \cite{chen2020lowresource}, as well as PIZZA~\cite{pizzaData}, a new semantic-parsing dataset in the domain of pizza ordering.

Prefix tuning was previously applied to two generation tasks~\cite{li2021prefixtuning}, summarization and table-to-text. In both of these, the decoded outputs are pure natural-language text sequences. By contrast, in semantic parsing the output sequences contain \emph{special tokens} representing the non-terminals of the semantic grammar of the domain at hand, often containing bracketing symbols such as parentheses, square brackets, colons, and so on, e.g., {\tt [IN:GET\_LOCATION} in Figure~\ref{fig:semparse}. This does not present a problem for full fine-tuning, because full fine-tuning updates the entire model during training, including the embedding layer, and hence the subwords corresponding to these special tokens can be modified as needed. However, this is not possible during prefix tuning, which freezes the entire PLM during training. Consequently, as shown by our results, a straightforward application of prefix tuning to semantic parsing will not work.

We solve this problem by adding the domain's special tokens to the vocabulary, expanding the embedding matrix accordingly, and then unfreezing only those matrix entries that correspond to special tokens. More specifically, we unfreeze the entire matrix but set the gradients to zero for those cells that do not involve special tokens. We demonstrate that this change has a dramatic impact on the performance of prefix tuning for semantic parsing, while adding only a trivial number of additional parameters to the overall budget. As a result, prefix tuning performance on semantic parsing tasks is competitive with regular fine-tuning in full data regimes and outperforms in low data settings.

The rest of the paper is structured as follows: in Sections~\ref{Sec:PrefixTuning} and~\ref{Sec:BitFit} we introduce the two lightweight tuning techniques we consider: prefix tuning and BitFit. We present the datasets and experimental setup in Section~\ref{Sec:Datasets}, and our full-data and few-shot results in Section~\ref{Sec:Results}. Section~\ref{Sec:Analysis} analyzes prefix tuning and its variants. Section~\ref{Sec:Related} reviews related work, and Section~\ref{Sec:Concl} presents our conclusions.

\section{Prefix Tuning}\label{Sec:PrefixTuning}

Following recent work by~\citet{li2021prefixtuning}, we use prefix tuning as an alternative to fine-tuning a PLM for conditional generation,
applying it to semantic parsing. Prefix tuning is inspired from in-context learning (aka \emph{prompting}) with very large PLMs such as GPT-3  \cite{brown2020language},
whereby desired outputs are generated without any task-specific fine-tuning. Instead, prompting freezes the PLM's parameters and attempts
to guide the model by prefacing the input sequence with a natural language text that essentially gives instructions on what output to generate,
along with a few (input, output) examples.  Choosing the right text for prompting is important for performance but can be challenging~\cite{tacl_a_00324, lester2021power}.

Prefix tuning builds on this idea by formulating prompting as an optimization task, one that optimizes a set of
task-specific continuous column vectors prepended to the input, or {\em prefixes}, instead of input-specific discrete natural language tokens.
This allows for more expressive prompt forms while limiting the number of trainable parameters to as few as 0.1\% of the PLM parameters.

We start with a pretrained transformer-based  encoder-decoder language model capturing a distribution $p_{\phi}(y|x)$ \cite{vaswani17, lewis2019bart},
where $\phi$ are the model's parameters, the input $x$ is encoded by a bidirectional encoder, and the target sequence $y$ is generated autoregressively
by the decoder, conditioned on the encoded input $x$ and previously generated outputs. We define $z=[x;y]$ as the concatenation of $x$ and $y$.

For the semantic parsing task, we prepend prefixes to both the encoder and decoder to obtain $z=[\mathit{PREFIX};x;\mathit{PREFIX'};y]$; 
see Figure~\ref{fig:models}.
The prefixes are prepended to the key and value of each attention layer in the transformer architecture, including the encoder self-attention, decoder self-attention,
and encoder-decoder cross attention. The prefix length $\mathit{PrefixLength}$ is a hyperparameter and all hidden-layer representations of the prefixes are optimized.
The prefix parameters are stored in a trainable matrix $P_\theta$ with the following number of parameters:
\begin{eqnarray}
  \temv{Size}(P_{\theta}) & = & \mbox{\emph{PrefixLength}} \times \temv{Size}(\temv{layers}) \nonumber \\
  {} & {} & \times \; \mathit{HidDim} \times 2 \times 3 \label{eq:prefixSize},
\end{eqnarray}
where $\mathit{HidDim}$ is the dimension of the hidden state in a single attention layer, the {\tt 2} multiplier is due to the attention key and value parameters, and the {\tt 3} multiplier is due to the three aforementioned types of attention layers. The model is then trained similarly to the original language model, but now only modifying the prefix parameters $\theta$, as the $\phi$ parameters remain frozen.  Furthermore, these prefix parameters are trained on task-specific training data.  Each semantic parsing domain is considered a different
task and results in a different set of task-specific prefix-tuning parameters.

\subsection{Special Token Embeddings}\label{Sec:SpecialTokens}

Prefix tuning was shown to give strong results in the summarization and table-to-text tasks \cite{li2021prefixtuning}, but its effectiveness on the semantic parsing task is not a priori clear. In both summarization and table-to-text, the output sequences are pure natural language. In a semantic parsing task, by contrast, the output semantic representations often\footnote{But not always; sometimes semantic parsing models are trained to output natural language directly, see Section~\ref{Sec:Related}.} contain special labels representing  non-terminal nodes, which are usually formulated by concatenating multiple tokens together with underscores, colons, brackets, and so on, such as {\tt [IN:GET\_REMINDER\_DATA\_TIME}. In prefix tuning, these special labels are tokenized by the PLM's subword tokenizer. The resulting tokenized sequences will deviate from the data given to the PLM during pretraining, making it much more difficult for the decoder to generate the correct output sequences; we show this in Section~\ref{Sec:Results}.

To address this problem, we introduce \emph{special token embeddings} to the prefix tuning method: We add the non-terminal labels as special tokens to the tokenizer vocabulary, expand the pretrained embedding layer to the size of the new vocabulary, and tune the embedding vectors of the special tokens, $E_\eta$, during training along with the prefixes, while keeping the embeddings of all other tokens fixed. With special token embeddings, the number of additional parameters that need to be trained and stored is
\begin{equation}
Size(E_{\eta}) = Size(labels) \times EmbDim \label{eq:specialEmbSize},
\end{equation}
where $\mathit{EmbDim}$ is the dimension of the embedding vectors and the number of labels is domain specific.

During training, instead of initializing the special token embeddings randomly, we use the averaged embedding vectors of their subword tokens with the original vocabulary as the initial values. Furthermore, we remove the open bracket from the label tokens and rely on the subword tokenization to split the open bracket from labels.

In Section~\ref{Sec:Results} we demonstrate that these special token embeddings significantly improve the performance of prefix tuning on semantic parsing, especially in the full-data setting.

\section{BitFit}
\label{Sec:BitFit}

BitFit~\cite{zaken2021bitfit}, or Bias-terms Fine-tuning, has recently been proposed as a method of fine-tuning an encoder-based PLM for downstream tasks, such as GLUE~\cite{wang2019glue} sentence classification tasks, by adapting only the bias parameters of the model. Specifically, for a pretrained transformer encoder containing linear components $\mathbf{W} \cdot \mathbf{x} + \mathbf{b}$, BitFit fine-tunes only the vector of bias terms $\mathbf{b}$. This is shown to achieve competitive performance with full fine-tuning and other lightweight fine-tuning methods on sentence and token-level classification tasks, especially in low resource settings. In addition, it is shown that the bias parameters are special in this sense, and that a random subset of parameters selected from the model do not achieve similar fine-tuning performance.

Here we apply BitFit to a new task, semantic parsing, which involves autoregressive generation and an encoder-decoder architecture, and compare its performance against other adaptation techniques. We extend the set of fine-tuned parameters to include the biases in the decoder. Thus, the set of fine-tuned parameters includes the biases of the query, key, and value projections in the encoder self-attention, the decoder self-attention, and the decoder cross-attention. In addition, we follow~\citet{zaken2021bitfit} and include the biases in the layer norm and feed-forward components from the encoder and decoder.

\section{Experimental Setup}\label{Sec:Datasets}

\subsection{Datasets}

We use two public datasets for task-oriented semantic parsing in our experiments:  the {\tt TOPv2} dataset \cite{chen2020lowresource}, and a recently open-sourced food-ordering dataset called {\tt PIZZA} \cite{pizzaData}.

The TOPv2 dataset contains 180K examples in 8 distinct domains: alarm, event, messaging, music, navigation, reminder, timer, and weather. 
Each example in this dataset contains three segments: a domain label, a natural language utterance, and a target semantic tree consisting of intent nodes, slot nodes, and leaves that contain fragments of the utterance text. The original semantic representations provided in the dataset are in the TOP format. In this paper we convert the TOP format to a TOP-Decoupled format \cite{conversationalTOP} by removing terminals that are not associated with any semantic nodes in the parsing tree.
For example, the source utterance in Figure~\ref{fig:semparse},
\emph{"What's happening in Harvard Square"},
\medskip
\noindent is mapped to the target semantics: \\[-0.18in]
\begin{center}
  \begin{tabular}{l}
    \mtt{[IN:GET\_EVENT} \\
    \hspace*{0.1in} \mtt{[SL:LOCATION} \\
    \hspace*{0.2in} \mtt{[IN:GET\_LOCATION} \\
    \hspace*{0.3in} \mtt{[SL:POINT\_ON\_MAP} \emph{Harvard Square}\mtt{ ] ] ] ]} \\[0.03in]
\end{tabular}
\end{center} \noindent
This representation differs from the TOP format in that tokens that do not carry any semantics are removed: \{\emph{What's happening in}\} do not appear in the above target. Only children representing resolvable entities of interest are kept.

We report results on three TOPv2 domains, selected so as to have diverse profiles in structural complexity (see table below):

\begin{itemize}
\itemsep0em
  \item {\tt Navigation} is the domain with the largest average tree depth and number of slots. 
  \item {\tt Reminder} has the second deepest semantics, and has the largest intent catalog of all 8 domains. 
   \item {\tt Weather} is on the other end of the spectrum: it is the domain with the smallest average depth and smallest catalog sizes. 
\end{itemize}
The following table lists the number of examples, intents, and slots, as well as the average semantic-tree depth for these three domains:  \\[-0.05in]
{\begin{center}
    \begin{tabular}{|l||c|c|c|}
        \hline
        & Weather & Navigation & Reminder \\    \hline
        \#Train    & 23,054  & 20,998  & 17,840  \\  \hline
        \#Dev      & 2,667   & 2971   & 2,526   \\  \hline
        \#Test     & 5,682   & 6,075   & 5,767   \\  \hline
        \#Intent   & 7      & 17     & 19     \\  \hline
        \#Slot     & 11     & 33     & 32     \\  \hline
        Depth      & 1.93   & 2.68   & 2.45   \\  \hline
    \end{tabular}
\end{center}} \noindent \\[-0.12in]

We also experiment with a low-data regime and use the \textbf{10SPIS} (samples per intent and slot) setting used by~\citet{chen2020lowresource} for the \emph{Reminder} and \emph{Weather} domains; relevant statistics can be found below. As no low-resource splits were provided for the \emph{Navigation} domain, we sampled its 10SPIS fold ourselves. \\[-0.03in]
{\begin{center}
		\begin{tabular}{|l||c|c|c|c|}
			\hline
			& Weather & Navigation & Reminder \\    \hline
			\#Train    & 87 & 156  & 219  \\  \hline
			\#Dev      & 59  & 132 & 161  \\  \hline
			\#Test     & 5,682 & 6,075  & 5,767   \\  \hline
		\end{tabular}
\end{center}} \noindent \\[-0.02in]
The test set is the same as for the full-data setting. 

The PIZZA dataset contains about 2 million synthetically generated pizza and/or drink orders mapped to their target semantics, used for training, and about 1.7K human-generated and annotated orders used for validation and testing. The dataset also comes in the TOP format. We follow the same rules used in the TOPv2 dataset to form the TOP-Decoupled semantics.

For the PIZZA dataset, we simulate a few-shot scenario as follows: Using the human-annotated validation set of 348 utterances, we randomly sample 200 utterances from it, and then incrementally build 30-, 50-, and 100-shot datasets. For each of these three datasets, we also use validation sets of sizes 30, 50, and 100, respectively. 
We report results on the provided test set, which contains 1,357 utterances. \\[0.004in]
	\begin{center}
		\begin{tabular}{|l||c|c|c|}
			\hline
			& 30 shots & 50 shots & 100 shots \\    \hline
			\#Train    & 30  & 50  & 100  \\  \hline
			\#Dev      & 30   & 50   & 100   \\  \hline
			\#Test     & 1,357   & 1,357   & 1,357   \\  \hline
		\end{tabular}
        \end{center} \noindent
        \\[-0.15in]

As a point of comparison, the average semantic tree depth of the human-generated portion of the {\tt PIZZA} dataset is around 3.6.

\subsection{Metric}

Even though the training data is in TOP-Decoupled format, the trained model might still produce terminals that are not associated with slot nodes, and which therefore do not contribute to the output's semantics; we remove such terminals for evaluation purposes. Moreover, since sibling order in a parse tree does not alter its semantics (in both datasets), we ignore ordering differences when computing the semantics-only exact-match metric. We refer to this as the Unordered Semantics-Only Exact-Match metric (EM). 

\subsection{Architectures and Hyperparameters}

The PLM we use in all of our experiments is BART-Large, a 24-layer transformer-based encoder-decoder model pretrained on a masked span prediction task on 160GB of unlabeled text \cite{lewis2019bart}. The pretraining task involves masking contiguous spans of an input sequence, encoding the masked sequence with the 12-layer encoder, and generating the original text using the 12-layer decoder. Therefore, the number of $\mathit{layers}$ in Equation~\ref{eq:prefixSize} is {\tt 12}. The dimensions of the hidden states, $\mathit{HidDim}$, and the embedding vectors, $\mathit{EmbDim}$, are both {\tt 1024}.

To study the effectiveness of different tuning strategies, we use full fine-tuning ({\tt FT}) as the baseline model. We then compare the accuracy of the two adaptation techniques: prefix tuning with and without special token embeddings
({\tt \textsuperscript{\textdagger}Prefix} and {\tt Prefix} respectively) and BitFit ({\tt BitFit}). We report main results with fixed prefix lengths of {\tt 30} and {\tt 5} for the full data setting, as well as {\tt 20} and {\tt 5} for the low data setting in the next section, and study the effect of prefix length on model accuracy more systematically in Section~\ref{Sec:Analysis}. We also include partial fine-tuning ({\tt FT-Top2}) in the comparison, which only fine-tunes the top 2 \emph{decoder} layers of the pretrained BART-Large model. The intuition behind this  approach is that the upper PLM layers contain more specialized information. 

To establish a fair comparison across various tuning strategies, we allocate an equal amount of trials (16) to each method. Instead of grid search, we select the tunable hyperparameters and ranges (as described
in the Appendix~\ref{app:hp})
for each method and then conduct a random search~\cite{bergstra12a}.
All other hyperparameters are fixed during the random search and their values are listed in Table~\ref{table:fixed_hp}.
Each trial is repeated over three random seeds and the hyperparameters with the best average performance on the validation set are selected for final evaluation.

\begin{center}
    \begin{table}[!hbt]
    {    \centering
    \begin{center}
        \begin{tabular}[c]{l|ll}
            \toprule
                hyperparameter & value \\
                \midrule
                beam size & 6 \\
                max epoch & 50 \\
                early stopping patience & 4 \\
                max target length & 200 \\
                gradient accumulation step & 3 \\
                evaluation frequency & 1 epoch* \\
            \bottomrule
        \end{tabular}
    \end{center}
    \caption{Values of the fixed hyperparameters in all experiments. *In low data settings, we duplicate the training data multiple times to have similar number of training examples as the full data settings (20,000 examples), so that the number of updates between evaluations are similar across all experiments.}
    \label{table:fixed_hp}
    }
    \end{table}
\end{center}

\section{Results}\label{Sec:Results}
\subsection{Full Data Setting}

Table~\ref{table:main_results} presents results for the full-data setting. We see that both prefix tuning ({\tt Prefix\textsubscript{len}}) and BitFit significantly underperform both full and partial fine-tuning. The biggest performance drop is as large as 20.72 absolute points. In the case of prefix tuning, this is likely because the target sequences are not natural language. Because the PLM is not fine-tuned, the decoder cannot learn anything about the rather unique non-terminal labels that occur in the outputs (and are unlikely to have appeared in the datasets used to pretrain the PLM). In the case of BitFit, we find that while tuning bias terms works well for sentence classification tasks, its performance degrades significantly on token-level tasks (we also tried BitFit for named entity recognition, in addition to semantic parsing) and tasks involving generation.

We solve the prefix tuning problem by adding the special labels to the tokenizer vocabulary and tuning the embedding vectors of these special tokens together along with the prefixes. Table~\ref{table:main_results} shows that with special token embeddings added in this way ({\tt \textsuperscript{\textdagger}Prefix\textsubscript{len}}), the exact match accuracy of prefix tuning can be boosted by 9.61 to 17.49 absolute points, with only a negligible number of additional task-specific parameters added (0.01\% of the total parameter size). Note that the number of task-specific parameters added by the special token embeddings is proportional to the number of non-terminal labels, including both intents and slots, in the dataset. The numbers reported in Table~\ref{table:main_results} are obtained from the {\tt Reminder} domain, which has the largest number of non-terminal labels in the TOPv2 dataset. Therefore, it represents an upper bound on the number of task-specific parameters for that entire dataset.

The results with prefix lengths of 30 and 5 suggest that in harder domains like {\tt Navigation}, longer prefixes are preferable, as they help to learn complex semantic structures. By contrast, in relatively easy domains like {\tt Weather}, shorter prefixes work better, as they avoid overfitting.

Compared to full fine-tuning, prefix tuning with special token embeddings can achieve competitive performance with only 0.1\% of the PLM's parameters. It also decidedly outperforms the BitFit method, with a similar number of parameters.

\begin{center}
    \begin{table*}[!t]
    {\small
    \centering
    \begin{center}
        \begin{tabular}[c]{l|ccc|c|cc}
            \toprule
                        & \multicolumn{4}{c|}{\textbf{TOPv2}}
                        & \multicolumn{2}{c}{\textbf{Task Specific Parameters}} \\
                        & Navigation & Reminder & Weather & Avg. & Number & \% of PLM Parameters \\
            \midrule
            FT          & 83.08 & 82.66 & 89.81 & 85.18 & 406,291,456 & 100\% \\%\vspace{2pt} \\
            \midrule
            FT-Top2     & 77.38 & 75.29 & 88.66 & 80.44 & 33,629,273 & 8.28\% \\
            Prefix\textsubscript{30}   & 63.05 & 65.61 & 72.75 & 67.14 & 2,211,840 & 0.54\% \\
            Prefix\textsubscript{5}    & 62.36 & 69.29 & 76.75 & 69.46 & 368,640 & 0.09\% \\
            \textsuperscript{\textdagger}Prefix\textsubscript{30}   & \textbf{80.41} & \textbf{79.12} & 87.49 & 82.34 & 2,264,064 & 0.56\% \\
            \textsuperscript{\textdagger}Prefix\textsubscript{5}    & 79.85 & 78.90 & \textbf{88.88} & \textbf{82.54} & 420,864 & 0.10\% \\
            BitFit      & 67.55 & 62.87 & 70.66 & 67.03 & 320,399 & 0.08\% \\
            \bottomrule
        \end{tabular}
    \end{center}
    \caption{EM accuracy of models trained with the full TOPv2 datasets averaged over 3 seeds (left), and the number and percentage of task specific parameters (right). Symbol \textsuperscript{\textdagger} indicates the special token embeddings.}
    \label{table:main_results}
    }
    \end{table*}
\end{center}

\subsection{Low Data Setting}

The few-shot results on both PIZZA and TOPv2, presented in Table~\ref{table:lowdata_results},  show that off-the-shelf prefix tuning outperforms all other tuning strategies, including full fine-tuning, partial fine-tuning, and prefix tuning with special token embeddings. On average, it outperforms full fine-tuning by 2.76 absolute points, while requiring a significantly smaller number of parameters.

Because models are more prone to overfitting in low-data regimes, we hypothesize that, with more parameters available to the model, full fine-tuning and prefix tuning with special token embeddings overfit the training data and produce worse results on test data. Furthermore, according to ~\citet{adapterTuningEffectiveness}, the lower layers of a PLM capture more generic features while upper layers capture more task-specific features. Therefore, as the lowest layer in a PLM, if the special token embeddings in the embedding layer come to overfit the training data, the model will not be able to generalize well to unseen data.

Overall, prefix tuning with special token embeddings in the full-data regime and without special token embeddings in the low-data regime significantly outperforms the other lightweight tuning strategies. It is competitive with the full fine-tuning baseline when there is plenty of training data and performs even better in few-shot scenarios.

\begin{center}
    \begin{table*}[!h]
    {\small
    \centering
    \begin{center}
        \begin{tabular}[c]{l|ccc|c|ccc|c}
            \toprule
                        & \multicolumn{4}{c|}{\textbf{TOPv2 10SPIS}}
                        & \multicolumn{4}{c}{\textbf{PIZZA}} \\
                        & Navigation & Reminder & Weather & Avg. & 30 Shots & 50 Shots & 100 Shots & Avg. \\
            \midrule
            FT          & 44.89 & 51.46 & 62.75 & 53.03 & 65.70 & 72.00 & 83.50 & 73.73 \\%\vspace{2pt} \\
            \midrule
            FT-Top2     & 10.25 & 17.87 & 26.95 & 18.36 & 24.48 & 42.82 & 60.22 & 42.51 \\
            Prefix\textsubscript{20}   & 46.75 & 56.41 & 68.15 & 57.10 & \textbf{67.99} & \textbf{71.93} & \textbf{85.55} & \textbf{75.16} \\
            Prefix\textsubscript{5}    & 45.24 & \textbf{56.98} & \textbf{69.19} & \textbf{57.13} & 66.72 & 71.53 & 83.37 & 73.88 \\
            \textsuperscript{\textdagger}Prefix\textsubscript{20}   & 46.02 & 52.71 & 56.21 & 51.65 & 62.58 & 69.39 & 82.74 & 71.57 \\
            \textsuperscript{\textdagger}Prefix\textsubscript{5}    & \textbf{47.50} & 55.32 & 63.86 & 55.56 & 64.39 & 71.24 & 82.91 & 72.84 \\
            BitFit      & 2.88 & 13.25 & 43.66 & 19.93 & 48.35 & 57.49 & 70.22 & 58.69 \\
            \bottomrule
        \end{tabular}
    \end{center}
    \caption{EM accuracy of few-shot learning on TOPv2 dataset with 10SPIS (left) and PIZZA dataset (right). Symbol \textsuperscript{\textdagger} indicates the special token embeddings.}
    \label{table:lowdata_results}
    }
    \end{table*}
\end{center}

\section{Analysis}
\label{Sec:Analysis}

\subsection{Prefix Length}
\begin{figure}[h!]
    \centering
    \includegraphics[width=0.48\textwidth]{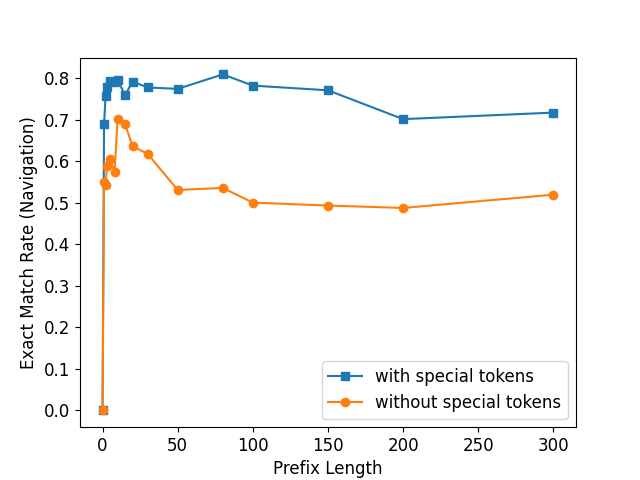}
    \caption{Effect of prefix length on exact match rate in Navigation domain.}
    \label{fig:plen_vs_em}
\end{figure}

To study the effect of prefix length on model accuracy, we trained prefix tuning models (with and without special token embeddings) in the Navigation domain with prefix lengths ranging from 0 to 300. The results are plotted in Figure~\ref{fig:plen_vs_em}. We see that as long as the prefix length is not zero, the models deliver decent accuracy even with prefix lengths as short as 1. Model performance increases as the prefix length increases, until it reaches an optimal length (80 with special token embeddings and 10 without). Then there is a performance drop, before the model finally attains a steady state.

The variance of model accuracy with special token embeddings is much smaller than that of the model without such embeddings, suggesting that the special token embeddings help to stabilize the model and make prefix tuning more robust with respect to prefix length. 

Figure~\ref{fig:plen_vs_em} also demonstrates that special token embeddings dramatically improve the prefix tuning performance across all the prefix lengths. The average performance boost is 19.98 absolute points.

\subsection{Prefix Variants}

In the prefix tuning model used to obtain the main results, we {\em prepend\/} prefixes to \emph{all attention layers} in both the encoder and the decoder. Here we explore two variants of this
technique: 1) inserting these vectors in different locations; and 2) prepending prefixes only in the top 2 decoder layers.

\paragraph{Different vector locations:}
Because BART's decoder is autoregressive and proceeds from left to right, prefixes are the only reasonable choice for where to add task-specific hidden vectors. However, encoder layers are bidirectional and symmetric, and therefore these vectors don't need to be prepended to the left of the attention layers. Accordingly, we investigate the effect of different location choices by:
\begin{itemize} \item \emph{appending} these vectors to the right of the encoder layers as suffixes ({\tt \textsuperscript{\textdagger}Suffix}); and by
\item splitting the prefix vectors into a prefix sequence and a suffix sequence ({\tt \textsuperscript{\textdagger}Prefix\&Suffix}).
\end{itemize}
  In the {\tt \textsuperscript{\textdagger}Prefix\&Suffix} model, the prefix vectors are split evenly with rounding up on the prefix side. For example, a {\tt \textsuperscript{\textdagger}Prefix\&Suffix\textsubscript{5}} model includes a prefix of length 3 and a suffix of length 2.

Table~\ref{table:prefix_location} shows that {\tt \textsuperscript{\textdagger}Suffix} models produce almost identical performance as the {\tt \textsuperscript{\textdagger}Prefix} models on average, while the \linebreak {\tt \textsuperscript{\textdagger}Prefix\&Suffix} model can slightly improve model performance by an average of 0.80 and 0.11 absolute points for prefix lengths of 30 and 5, respectively. This demonstrates that prefix tuning is not sensitive to prefix location.

\paragraph{Prefix tuning only the top 2 decoder layers:}
Inspired by the fine-tuning of the top 2 decoder layers, we study {\tt Prefix\textsuperscript{top-2-layers}}, a prefix-tuning variant that only prepends prefixes to the top two decoder layers. An additional motivation behind this experiment is that even though prefix tuning can dramatically reduce the number of task-specific parameters, it cannot reduce training latency, since the gradients still need to back-propagate to the very first layer in the encoder. On the other hand, with prefixes in only the top 2 decoder layers, the backward propagation can be terminated immediately after the lowest prefix has been processed, which should make the training process significantly faster.

The last row in Table~\ref{table:prefix_location} shows the resulting model accuracy when we only add prefixes to the top 2 decoder layers. We report the results we obtain from the best prefix length in each domain instead of using a fixed prefix length. As can be seen, performance drops radically for partial prefix tuning. EM accuracy regresses by 26.92 absolute points on average compared to full prefix tuning without special tokens, and by more than 40 absolute points compared to full prefix tuning with special tokens. 

\begin{center}
    \begin{table}[]
    {\small
    \centering
    \begin{center}
        \begin{tabular}[c]{l|ccc|c}
            \toprule
                        & \multicolumn{4}{c}{\textbf{TOPv2}} \\
                        & Navigation & Reminder & Weather & Avg. \\
            \midrule
            Prefix\textsubscript{30}   & 63.05 & 65.61 & 72.75 & 67.14 \\
            Prefix\textsubscript{5}    & 62.36 & 69.29 & 76.75 & 69.46 \\
            \textsuperscript{\textdagger}Prefix\textsubscript{30}   & 80.41 & 79.12 & 87.49 & 82.34 \\
            \textsuperscript{\textdagger}Prefix\textsubscript{5}    & 79.85 & 78.90 & 88.88 & 82.54 \\
            \midrule
            \textsuperscript{\textdagger}Suffix\textsubscript{30}   & 77.77 & 78.63 & \textbf{89.90} & 82.10 \\
            \textsuperscript{\textdagger}Suffix\textsubscript{5}    & 76.96 & 80.01 & 89.42 & 82.13 \\
            \textsuperscript{\textdagger}Prefix\&Suffix\textsubscript{30}   & 80.13 & \textbf{80.18} & 89.12 & \textbf{83.14} \\
            \textsuperscript{\textdagger}Prefix\&Suffix\textsubscript{5}    & \textbf{80.35} & 78.65 & 88.96 & 82.65 \\
            \midrule
            Prefix\textsuperscript{top-2-layers}   & 33.30 & 35.28 & 55.56 & 41.38 \\
            \bottomrule
        \end{tabular}
    \end{center}
    \caption{EM accuracy of different prefix tuning variants. Symbol \textsuperscript{\textdagger} indicates the special token embeddings.}
    \label{table:prefix_location}
    }
    \end{table}
\end{center}

\subsection{Effect of Special Token Embeddings}

The results in Section~\ref{Sec:Results} demonstrate that special token embeddings can dramatically improve the performance of prefix tuning in semantic parsing tasks. By adding embedding vectors for non-terminal labels, a frozen PLM with tunable prefixes {\em and\/} tunable special token embeddings can learn the distributed representations of the latter, which help to guide the decoder in the right directions.

Besides this contribution, we hypothesize that special token embeddings can also reduce the target sequence length and further improve model performance. In a transformer-based PLM such as BART, both source and target sequences are tokenized by the PLM's \emph{subword} tokenizer. Since the non-terminal labels in semantic representations are created artificially and tend to contain multiple words and special characters, they are usually tokenized to long subword sequences. For example, a typical intent label, {\tt IN:GET\_REMINDER\_DATA\_TIME}, in the Reminder domain, is tokenized to 12 subwords: {\tt ["IN", ":", "GET", "\_", "REM", "IND", "ER", "\_", "D", "ATE", "\_", "TIME"]}. Therefore, these special non-terminal labels cause the target sequences to be unnecessarily long, thereby complicating the decoding process. 

\begin{figure}[h!]
    \centering
    \includegraphics[width=0.48\textwidth]{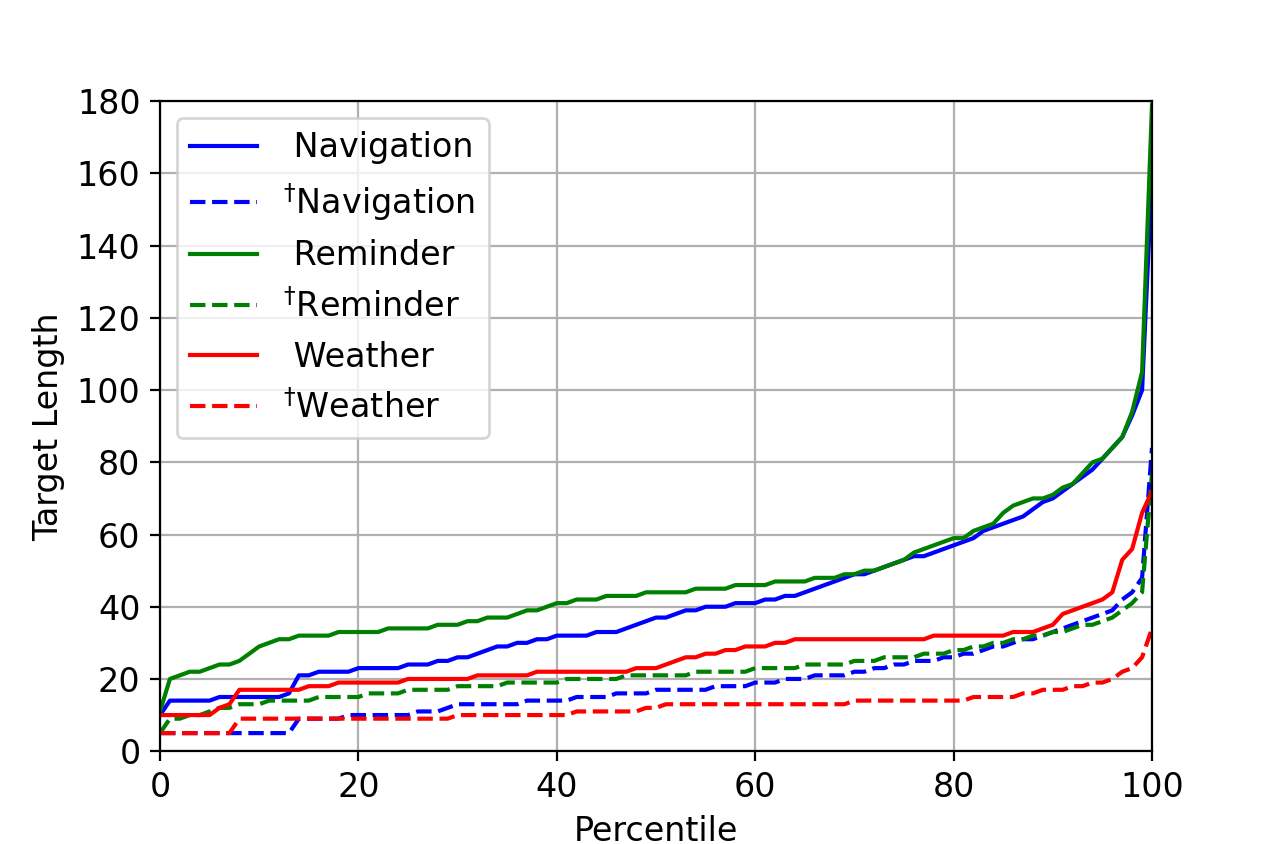}
    \caption{Percentile of target length in Navigation (blue), Reminder (green), and Weather (red) domains with (dash lines) and without (solid lines) special tokens. Symbol \textsuperscript{\textdagger} indicates the special token embeddings.}
    \label{fig:length_percentile}
\end{figure}

The issue can be mitigated with special token embeddings. By adding the non-terminal labels to the tokenizer subword vocabulary, these labels will stay as whole units after tokenization, which will reduce the overall target sequence length. Table~\ref{table:target_length} and Figure~\ref{fig:length_percentile} show the statistics and percentiles, respectively, of the target sequence lengths in the Navigation, Reminder, and Weather domains.
We can clearly see that special token embeddings significantly reduce target length. The mean target length is reduced by 53.67\% on average across three domains. With shorter target sequences, the decoder should be able to generate correct outputs easier.

To evaluate the effect of target sequence length on overall performance, we compare the full fine-tuning models with and without special token embeddings. Table~\ref{table:ft_special_tokens} shows that special token embeddings consistently boost model performance in all three domains. The average performance improvement is 1.09 absolute points. In full fine-tuning, the entire embedding layer, including the embedding vectors of the subwords corresponding to the non-terminal labels, is updated. Thus, the distributed representations of the non-terminal labels can be learned regardless of special token embeddings, which indicates that in full fine-tuning the performance improvement introduced by special token embeddings comes primarily from the reduced length of the target sequences. 

Therefore, adding special token embeddings to prefix tuning method helps not only to learn the distributed representations of non-terminal labels in a given semantic parsing task, but also to facilitate the decoding process by reducing overall target sequence length.

\begin{center}
    \begin{table}[]
    {\small
    \centering
    \begin{center}
        \begin{tabular}[c]{l|cccc}
            \toprule
            Domain & Max. & Min. & Mean & Median \\
            \midrule
            Navigation   & 160 & 10 & 40.41 & 37 \\
            \textsuperscript{\textdagger}Navigation  & 84 & 5 & 18.34 & 17 \\
            \midrule
            Reminder   & 179 & 11 & 46.52 & 44 \\
            \textsuperscript{\textdagger}Reminder  & 77 & 5 & 22.00 & 21 \\
            \midrule
            Weather   & 72 & 10 & 26.13 & 23 \\
            \textsuperscript{\textdagger}Weather  & 34 & 5 & 12.10 & 12 \\
            \bottomrule
        \end{tabular}
    \end{center}
    \caption{Statistics of target length in Navigation, Reminder, and Weather domain. Symbol \textsuperscript{\textdagger} indicates the special token embeddings.}
    \label{table:target_length}
    }
    \end{table}
\end{center}

\begin{center}
    \begin{table}[]
    {\small
    \centering
    \begin{center}
        \begin{tabular}[c]{l|ccc|c}
            \toprule
                        & \multicolumn{4}{c}{\textbf{TOPv2}} \\
                        & Navigation & Reminder & Weather & Avg. \\
            \midrule
            FT          & 83.08 & 82.66 & 89.81 & 85.18 \\
            \textsuperscript{\textdagger}FT   & \textbf{83.17} & \textbf{84.33} & \textbf{91.33} & \textbf{86.27} \\
            \bottomrule
        \end{tabular}
    \end{center}
    \caption{EM accuracy of fine-tuned models with (bottom) and without (top) special token embeddings. Symbol \textsuperscript{\textdagger} indicates the special token embeddings.}
    \label{table:ft_special_tokens}
    }
    \end{table}
\end{center}

\section{Related Work}\label{Sec:Related}

Semantic parsing has been a staple of NLP for decades, particularly in connection with question answering \cite{ZelleM96,ZettlemoyerC05}.
More recently, semantic parsing has received increased attention due to its relevance to digital voice assistants. That is especially the case 
for variants such as task-oriented parsing (which is the category of the {\tt TOPv2} dataset). 

As with other NLP tasks, SOTA results in semantic parsing, including task-oriented parsing, have been obtained by fine-tuning
large pretrained language models \cite{rongali2020don,liu2019roberta,chen2020lowresource}, particularly pretrained encoder-decoder
architectures such as BART \cite{lewis2019bart}. \citet{ConstrainedLMs} leverages the PLM's ability to generate
natural language by reframing semantic parsing as a paraphrasing task. They use the PLM to map utterances
into a canonical (but highly restricted) natural-language representation that can then be automatically mapped to
the final output target.

PLMs also help with bootstrapping a task-oriented parser from a small amount of training data. ~\citet{chen2020lowresource}
approach this by fine-tuning the PLM with other domain data and also by utilizing meta-learning techniques. ~\citet{tran-tan-2020-generating} tackle the problem
by using a PLM to generate synthetic training data.

The issue with fine-tuning large pretrained models is that doing so requires a lot of space, and this becomes infeasible when there are many downstream tasks. 
As we saw, transparent non-mutative methods such as prefix tuning \cite{li2021prefixtuning} adapt a large PLM in parameter-efficient ways.  Other approaches
add task-specific layers, or {\em adapters}, between layers of the PLM~\cite{Rebuffi17, pmlr-v97-houlsby19a, pfeiffer-etal-2021-adapterfusion}.
These adapter approaches are not as space-efficient, as they generally require up to 30 times more tunable parameters in order 
to achieve performance comparable to prefix tuning \cite{li2021prefixtuning}.
There are also opaque non-mutative techniques, such as side-tuning \cite{zhang2020sidetuning}, which further facilitate sharing across multiple tasks. 

\section{Conclusions}\label{Sec:Concl}

We studied a transparent non-mutative fine-tuning strategy, prefix tuning, and a lightweight mutative fine-tuning strategy, BitFit, specifically for semantic parsing tasks, and compared their performance against each other and also against full and partial fine-tuning. We observed that both BitFit and prefix tuning perform poorly in regular data settings, and BitFit also performs poorly in few-shot settings. To solve the problem with prefix tuning, we proposed the addition of special-token embeddings and  demonstrated that this can dramatically improve the technique's performance while adding only a trivial number of task-specific parameters. With the help of special token embeddings, prefix tuning becomes competitive with full fine-tuning in the full data setting and outperforms full fine-tuning in the low data regime, despite modifying only 0.1\% of the total number of PLM parameters.

\begin{acks}
The authors thank Xiang Lisa Li for helpful discussions on prefix tuning.
\end{acks}

\bibliographystyle{ACM-Reference-Format}
\bibliography{custom}

\appendix
\section{Hyperparameter Tuning}
\label{app:hp}

In prefix tuning without a fixed prefix length, we tune the learning rate ({\tt lr}), batch size ({\tt bsz}), prefix length ({\tt prefix\_length}), and the intermediate dimension of prefixes ({\tt mid\_dim}). The detailed values of these hyperparameters in each random search trial is listed in Table~\ref{table:hp_prefixtune}. In the prefix tuning experiments with fixed prefix length, we overwrite the {\tt prefix\_length} value in the trial by the fixed length.

In other tuning strategies, including full fine-tuning, partial fine-tuning, and BitFit, we only tune the learning rate ({\tt lr}) and batch size ({\tt bsz}), as specified in Table~\ref{table:hp_others}.

\begin{center}
    \begin{table}[H]
    {
    \centering
    \begin{center}
        \begin{tabular}[c]{l|llll}
            \toprule
                trial & lr & bsz & prefix\_length & mid\_dim \\
                \midrule
                0 & 3.19E-05 & 16 & 50 & 800 \\
                1 & 1.13E-04 & 4 & 50 & 300 \\
                2 & 3.00E-04 & 8 & 100 & 800 \\
                3 & 4.62E-05 & 16 & 30 & 600 \\
                4 & 4.27E-04 & 4 & 50 & 300 \\
                5 & 1.79E-05 & 16 & 30 & 300 \\
                6 & 5.89E-05 & 4 & 50 & 800 \\
                7 & 3.88E-05 & 16 & 20 & 600 \\
                8 & 5.93E-06 & 16 & 30 & 300 \\
                9 & 6.23E-05 & 16 & 100 & 300 \\
                10 & 5.81E-06 & 8 & 100 & 600 \\
                11 & 1.25E-04 & 4 & 50 & 800 \\
                12 & 2.64E-05 & 16 & 20 & 600 \\
                13 & 3.98E-05 & 4 & 20 & 300 \\
                14 & 1.25E-05 & 4 & 100 & 800 \\
                15 & 3.70E-05 & 16 & 20 & 300 \\
            \bottomrule
        \end{tabular}
    \end{center}
    \caption{Hyperparameter values in each random search trial for prefix tuning.}
    \label{table:hp_prefixtune}
    }
    \end{table}
\end{center}

\begin{center}
    \begin{table}[H]
    {
    \centering
    \begin{center}
        \begin{tabular}[c]{l|ll}
            \toprule
                trial & lr & bsz \\
                \midrule
                0 & 1.61E-04 & 4 \\
                1 & 7.34E-05 & 16 \\
                2 & 2.52E-04 & 4 \\
                3 & 1.22E-04 & 4 \\
                4 & 1.49E-05 & 8 \\
                5 & 3.69E-04 & 8 \\
                6 & 5.82E-05 & 4 \\
                7 & 1.50E-05 & 8 \\
                8 & 4.07E-05 & 16 \\
                9 & 1.40E-05 & 16 \\
                10 & 5.40E-06 & 16 \\
                11 & 1.50E-05 & 8 \\
                12 & 9.82E-05 & 4 \\
                13 & 7.07E-05 & 4 \\
                14 & 1.42E-04 & 8 \\
                15 & 4.38E-05 & 4 \\
            \bottomrule 
        \end{tabular}
    \end{center}
    \caption{Hyperparameter values in each random search trial for full fine-tuning, partial fine-tuning, and BitFit.}
    \label{table:hp_others}
    }
    \end{table}
\end{center}

\end{document}